\documentclass{article}

\usepackage[preprint]{neurips_2025}

\usepackage[utf8]{inputenc} 
\usepackage[T1]{fontenc}    
\usepackage{hyperref}       
\usepackage{url}            
\usepackage{booktabs}       
\usepackage{amsfonts}       
\usepackage{nicefrac}       
\usepackage{microtype}      
\usepackage{xcolor}         

\usepackage{amsmath,amssymb,amsthm}
\usepackage{siunitx}
\usepackage{todonotes}     
\usepackage[utf8]{inputenc} 
\usepackage[T1]{fontenc}    
\usepackage{hyperref}       
\usepackage{url}            
\usepackage{booktabs}       
\usepackage{amsfonts}       
\usepackage{nicefrac}       
\usepackage{microtype}      
\usepackage{xcolor}         
\usepackage{enumitem} 
\usepackage{amsmath,amssymb,amsthm}
\usepackage{siunitx}
\usepackage{graphicx}
\usepackage{subcaption}

\usepackage{algorithm}
\usepackage{algorithmic}
\usepackage{multirow}
\usepackage{makecell}

\title{Self-supervised Pretraining for Integrated Prediction and Planning of Automated Vehicles}

\author{
Yangang Ren\textsuperscript{\rm 1,2}\thanks{Co-first authors.},
Guojian Zhan$^{2*}$,
Chen Lv$^{3}$,
Jun Li\textsuperscript{\rm 1},
Fenghua Liang\textsuperscript{\rm 1$\dagger$},
Keqiang Li\textsuperscript{\rm 2}\thanks{Co-advisors. liangfh@changan.com.cn;likq@tsinghua.edu.cn.}\\
\vspace{0.1cm}
\textsuperscript{\rm 1} Changan Automobile
\textsuperscript{\rm 2} Tsinghua University
\textsuperscript{\rm 3} Nanyang Technological University\\
}

\begin{document}

\maketitle

\begin{abstract}
Predicting the future of surrounding agents and accordingly planning a safe, goal-directed trajectory are crucial for automated vehicles. 
Current methods typically rely on imitation learning to optimize metrics against the ground truth, often overlooking how scene understanding could enable more holistic trajectories. In this paper, we propose Plan-MAE, a unified pretraining framework for prediction and planning that capitalizes on masked autoencoders. Plan-MAE fuses critical contextual understanding via three dedicated tasks: reconstructing masked road networks to learn spatial correlations, agent trajectories to model social interactions, and navigation routes to capture destination intents. To further align vehicle dynamics and safety constraints, we incorporate a local sub-planning task predicting the ego-vehicle's near-term trajectory segment conditioned on earlier segment. This pretrained model is subsequently fine-tuned on downstream tasks to jointly generate the prediction and planning trajectories. Experiments on large-scale datasets demonstrate that Plan-MAE outperforms current methods on the planning metrics by a large margin and can serve as an important pre-training step for learning-based motion planner.
\end{abstract}

\section{Introduction}
Generating safe, comfortable and efficient trajectories is the critical capability and ultimate goal of autonomous driving~\cite{hu2023planning}. Achieving this objective remains challenging today, as it requires understanding the road topology, predicting the behaviors of surrounding agents, adhering to vehicle dynamics, and conforming to human-specified driving goals (typically conveyed via navigation routes). For this purpose, modeling and understanding the intrinsic relationships between these input modalities is a prerequisite to achieve human-level driving.

Early research typically treats trajectory planning as an independent module, which takes the perceived and predicted vectors as inputs and is trained by imitation learning on the annotated data~\cite{vitelli2022safetynet, ren2025master}. Most of these planning models follow the success of learning-based prediction models and inherently incorporate the historical trajectory of the autonomous vehicle as input features. Despite rapid progress enabled by imitation-based planners, isolated optimization for planning sometimes fails to capture the interactions between surrounding agents and the ego vehicle, thereby limiting their generalization capability in dynamic scenarios~\cite{hagedorn2024integration}. Recent methods increasingly integrate prediction and planning in a joint or interdependent step to model bidirectional interactions~\cite{huang2023differentiable, huang2023gameformer,liu2025hybrid}. 
These models utilize the shared context to jointly output the prediction and planning trajectories, and some of them further use the prediction outcomes from the previous level to iteratively refine the trajectories of current level. Although demonstrated SOTA performance in the open-source benchmark such as nuPlan~\cite{karnchanachari2024towards}, these approaches markedly increase the complexity of the model architecture and the scale of trainable parameters. Plus, the pure imitation learning in these models tends to exploit any surface-level correlations that rapidly minimize the supervised loss, thereby learning shortcuts that appear correct but lack causal grounding~\cite{wen2020fighting,cheng2024rethinking}.

To address this challenge, we aim to incorporate the self-supervised learning to enhance the model’s ability to capture interactions among the input contexts, serving as a pretraining step ahead of the downstream imitation learning. Notably, the pretraining on images and LiDAR point clouds has been widely adopted in the perception module, substantially reducing reliance on labeled data~\cite{yang2024unipad, Hess_2023_WACV}. And the pretraining based on masking and reconstruction has been shown to mitigate overfitting to annotated labels in the prediction domain~\cite{chen2023traj,cheng2023forecast,lan2023sept}.
Driven by these observations, in this paper we propose Plan-MAE, a unified pretraining framework for integrated prediction and planning that uses Masked Autoencoders (MAE)~\cite{he2022masked} to master the scene understanding ability. Specifically, Plan-MAE consists of three auxiliary tasks that focus on the principal aspects of driving scenes:
(1) the masking and reconstruction of surrounding agents’ trajectories to capture social interactions; (2) the masking and reconstruction of map lanes to encode road constraints;
(3) the prediction of the navigation to learn the ego vehicle’s where-to-go information. And it also includes an alignment task where the ego vehicle’s historical trajectory is partitioned into head and tail segments, after which the head is utilized to predict the tail by leveraging the embeddings from the auxiliary tasks. This sub-planning helps the encoder establish a comprehensive understanding of the three types of unimodal contexts, meanwhile learning the vehicle dynamics and safe driving maneuvers from human trajectories.

We evaluate Plan-MAE on large-scale self-collected driving datasets labeled with expert trajectories. Compared with existing methods that likewise follow the imitation learning framework, our approach achieves substantial improvements in the planning accuracy, demonstrating that the pretraining is helpful to learn effective scene representations. Compared to existing prediction‐based pretraining methods, our planning-oriented pretraining consistently enhances planning performance through the collaboration of auxiliary and align tasks, even though this coherence does not extend to the predictions.

\section{Related Work}
\paragraph{Learning-based motion planning.}
Benefiting from the data‐driven nature, learning‐based planning has sparked widespread attention in autonomous driving, and is regarded as a scalable solution to complex scenarios~\cite{chen2024end}. Among these, imitation-based planners that supervise the model learning with expert trajectories, are widely adopted because of their ease of training and typical scalability with data. In the early stage, end-to-end approaches are proposed to map raw inputs directly to future trajectories~\cite{chitta2021neat,hu2022st}. However, this approach substantially increases the demand for labeled data, thereby necessitating reliance on open-source simulation platforms like Carla~\cite{dosovitskiy2017carla} for data collection. Another more effective approach is to leverage the post-perception outcomes as inputs and to directly train from recorded real-world data~\cite{jiang2023vad}. Works such as VectorNet~\cite{gao2020vectornet} and DenseTNT~\cite{gu2021densetnt} pioneered the use of vectorized scene representation, demonstrating a large-margin improvement in trajectory predictions. Inspired by this, planning-oriented models such as SafetyNet~\cite{vitelli2022safetynet} and UrbanDriven~\cite{scheel2022urban} have been proposed to replace traditional rule-based methods, posing a remarkable advance in real-world scenarios. Recently, the integrated prediction and planning has been proven as a more powerful approach to an imitative planner, owing to its modeling of interactions through joint trajectory generation~\cite{huang2023differentiable,liu2025hybrid, pini2023safe}. Combined with the data augmentation techniques such as state perturbation and dropout, this kind of models has achieved state-of-the-art in the nuPlan dataset and simulator\cite{cheng2024rethinking, cheng2024pluto}.

\paragraph{Self-supervised learning in trajectory prediction.}
Inspired by the masked autoencoders (MAE)\cite{he2022masked}, several recent studies have explored self-supervised learning for motion forecasting, aiming to predict the future trajectories of surrounding agents. Traj-MAE~\cite{chen2023traj} formulates two separate reconstruction tasks for agent trajectories and road maps, training their respective encoders independently under a self-supervised learning paradigm. To better capture temporal dependencies, Forecast-MAE~\cite{cheng2023forecast} further introduces a bidirectional prediction on the trajectories, where either its history or future is masked and then recovered from the remaining portion. This formulation is proven to encourage the trajectory encoder to extract common representations of both past and future dynamics. However, both Traj-MAE and Forecast-MAE adopt unimodal pretraining strategies, potentially underutilizing the cross-modal relationships between agents and maps.
SEPT~\cite{lan2023sept} bridges this gap by aligning trajectory and map embeddings within a sub-prediction task. For a given agent, the tail segment of its historical trajectory is predicted using the head segment, together with the features shared by the two preceding tasks. While this method offers valuable insights for our work, it remains focusing on trajectory prediction, and its pretraining benefits for enhancing planning performance have yet to be established.

Specially, motion planning not only requires accurate predictions of surrounding agents, but also generates reasonable trajectories that comply with navigation and vehicle dynamics. Moreover, recent studies have shown that improvements in prediction accuracy do not necessarily contribute to better planning performance, highlighting the need for planning-oriented representations \cite{hu2023planning, wang2024driving}.
In contrast to prior works, our proposed Plan-MAE extends the masked autoencoder paradigm to the motion planning domain, enhancing the model’s understanding of diverse environmental entities through masking and reconstruction tasks. By joint designing auxiliary tasks for scene understanding and their alignment to the ego's trajectory, Plan-MAE simultaneously learns the spatiotemporal representations and goal-aware ability, making it the first pretraining framework tailored for integrated prediction and planning in autonomous driving.

\section{Method}
\label{sec:method}

\subsection{Input Representations}
Our model uses the vector-based features as inputs, consisting of four components: current ego state, agent trajectory, map lanes, and the navigation route. All features are transformed into the ego-centric coordinate system, with the ego vehicle's current position as the origin.

\textbf{Ego state} is represented as a single-frame tensor with the shape of $[1, D_e]$, which contains attributes of the automated vehicle at the current frame. Here, we deliberately exclude the ego vehicle's historical trajectory from model inputs to avoid the well-established "copycat" problem~\cite{wen2020fighting} or learning shortcuts~\cite{geirhos2020shortcut}, where the model relies on trajectory extrapolation without fully understanding the underlying causal relationships. $D_e$ is the dimension of attribute, including the longitudinal and lateral positions, heading, longitudinal and lateral velocities, accelerations, and constant attributes such as the length, width and height.

\textbf{Agent trajectory} is extracted as a spatiotemporal tensor with the shape of $[N, T_h+1, D_n]$, which spatially consists of the attribute of $N$ nearest surrounding agents, and temporally the past $T_h$ plus the current frame of trajectories. $D_n$ is the attribute dimension of any agent, which records the same kinematic and shape features as the ego vehicle, augmented with the additional position offsets in the agent's local coordinates at the current timestamp. Considering that some of these agents may not have full $T_h$ frames of history, we add zero-padding to the short trajectory and properly mask them in the attention computation.

\textbf{Map lanes} capture the road geometry around the ego and target agents, represented as a tensor shaped of $[M, P_m, D_m]$. They encompass the $M$ nearest lanes relative to the ego vehicle at the current timestamp, with each lane represented by $P_m$ uniformly sampled centerline points spaced at 0.5-meter intervals. And each point has $D_m$ attributes such as positions, heading, type, color, left and right lane boundaries. All lanes are retrieved by searching within a radius of 200 m around the ego vehicle, and lanes that are too short are padded with zeros.

\textbf{Navigation route} is extracted as the tensor shaped of $[P_r, D_r]$ to indicate the ego's where-to-go information. It is derived by backtracking from the ego vehicle’s future trajectory and contains $P_r$ points to guide the ego vehicle’s position. For each point, the dimension $D_r$ contains the expected positions and heading. This curve always originates near the ego vehicle and extends forward up to 1500 meters.
\begin{figure*}[thbp]
\centering
\includegraphics[width=1.0\textwidth]{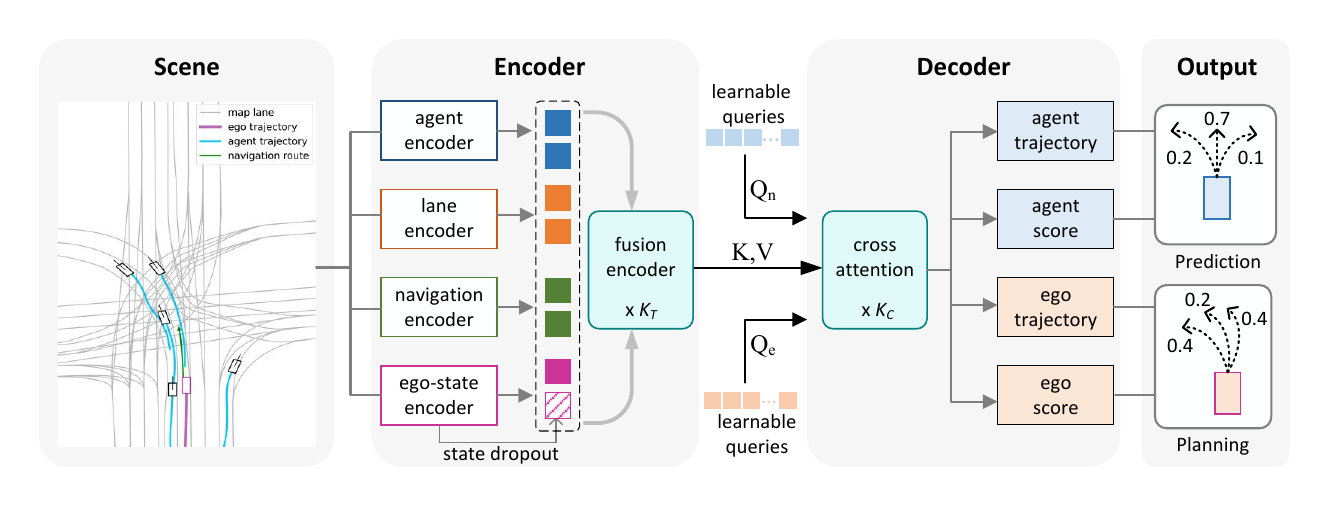}
\caption{Model for integrated prediction and planning.}
\label{fig_overview}
\end{figure*}

\subsection{Model for Integrated Prediction and Planning}

Building upon the distinct traffic elements, we construct a transformer-based model for the integrated prediction and planning, as depicted in Figure \ref{fig_overview}. This model includes an encoder to understand the relationships between multiple scene elements, a decoder to output the multi-modal trajectories and scores of the ego and agents. We use a game-theoretic method like GameFormer\cite{huang2023gameformer} to process the interaction between prediction and planning, wherein the outputs from the last level will be fused with the shared environmental context to generate the next-level trajectories.

\textbf{Agent encoder} firstly projects agents' trajectories into a high-dimension embedding space $\mathbb{R}^D$, outputting the embeddings $[N, T_h+1, D]$. The project layers are constructed with two linear layers with ReLU activation. After that, the embeddings across time steps are extracted through max pooling to produce the aggregated embeddings, shaped of $[N, D]$.

\textbf{Ego-state encoder} adopts dropout to prevent shortcut learning in kinematic states. Referenced the PlanTF~\cite{cheng2024rethinking}, we pick 6 variables from the ego's current frame, i.e., positions, heading, velocity, acceleration and steer, as the input. Each variable is equipped with a linear layer to generate embedding, and then combined with position encoding. During training, each embedded state token, except positions and heading, will be randomly dropped with a probability of 0.75. Ultimately, the ego state is processed into a $[1, D]$ embedding.

\textbf{Map and navigation encoder} share the same architecture. First, their original inputs are projected into a unified high-dimension feature space $\mathbb{R}^D$ using the two-layer MLP with ReLU activations. Then, a max pooling is conducted along the point sequence of each lane to aggregate spatial features. As a result, the features of map lanes and navigation route are processed into the embeddings shaped of $[M, D]$ and $[1, D]$, respectively. These embeddings are further concatenated as the embedding $[M+1, D]$ to formulate a comprehensive representation of road work.

\textbf{Fusion encoder} concatenates the ego features, map features and agents features to formulate a scene context embedding shaped of $[1+N+M+1, D]$. After that, a Transformer encoder with $K_T$ blocks is used to capture the relationships among all the scene elements. Each block incorporates a multi-head self-attention operation, with zero-padded elements (e.g., lanes and agents) being masked during attention computation.

\textbf{Decoder} is designed to predict the future $T_f$-step trajectories from the latent representations. We initialize $m$ learnable embeddings serving as the query, which will guide the generation of $m$ possible trajectories for each agent. Then, the $K_C$ multi-head cross-attention layers are utilized to fuse the initial modalities and the scene context from the encoder. Finally, the resulting query contents from cross-attention are delivered separately into the prediction and planning head to decode the agent’s trajectories along with their confidences. To improve the stability of long-horizon predictions, we employ an autoregressive approach to decode the trajectory point sequentially. At one step, the decoder is provided with the sequence it has generated thus far and is tasked with predicting the next point. Consequently, the model outputs the ego vehicle’s $[m, T_f, 2]$ trajectory with $m$ scores, and the agents’ $[N, m, T_f, 2]$ trajectory with $[N,m]$ scores.

\subsection{Pretrain for scene representation}
\begin{figure}[th]
  \centering
\includegraphics[width=0.62\textwidth]{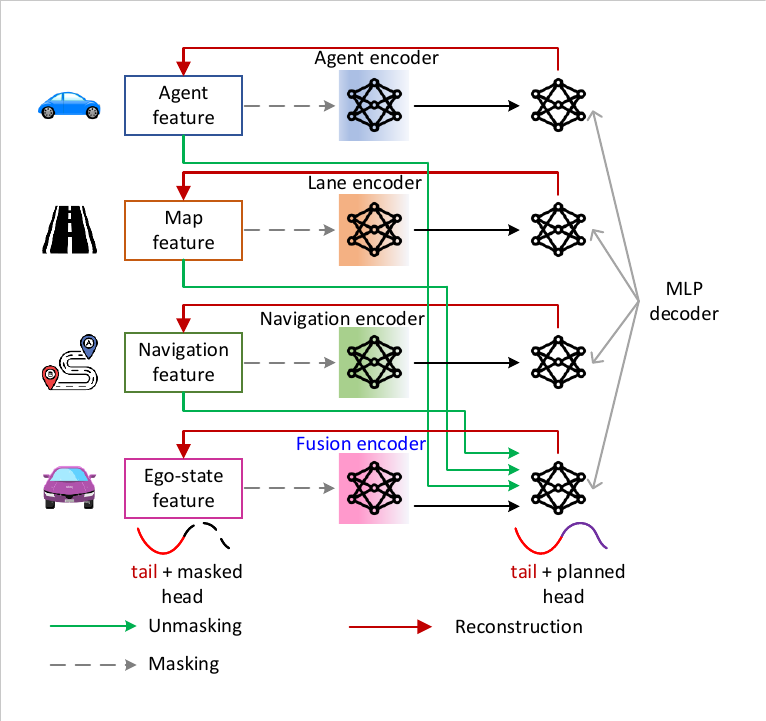}
  \caption{Framework of Plan-MAE.}
  \label{fig_pretrain}
\end{figure}

To understand the spatiotemporal relationships of driving scenarios, we design a self-supervised manner to capture the significant features for the ultimate planning. As shown in Figure~\ref{fig_pretrain}, three auxiliary tasks concerning agent trajectory, map lane, and navigation route are designed respectively to train their corresponding encoders. For each of them, a subset of input tokens is masked using zero-padding, and the encoder, along with a two-layer MLP, is trained to recover the original contents. Then, a local planning task on the ego's trajectory is utilized to train the fusion encoder, which combines the unmasked input features to predict the head trajectory based on the tail counterpart. Finally, the losses from these tasks are summed together as the optimization objective of the pretraining stage.

\begin{figure}[ht]
  \centering
  \begin{subfigure}[b]{0.22\textwidth}
\includegraphics[width=\textwidth,trim={0cm 0cm 0cm 0cm},clip]{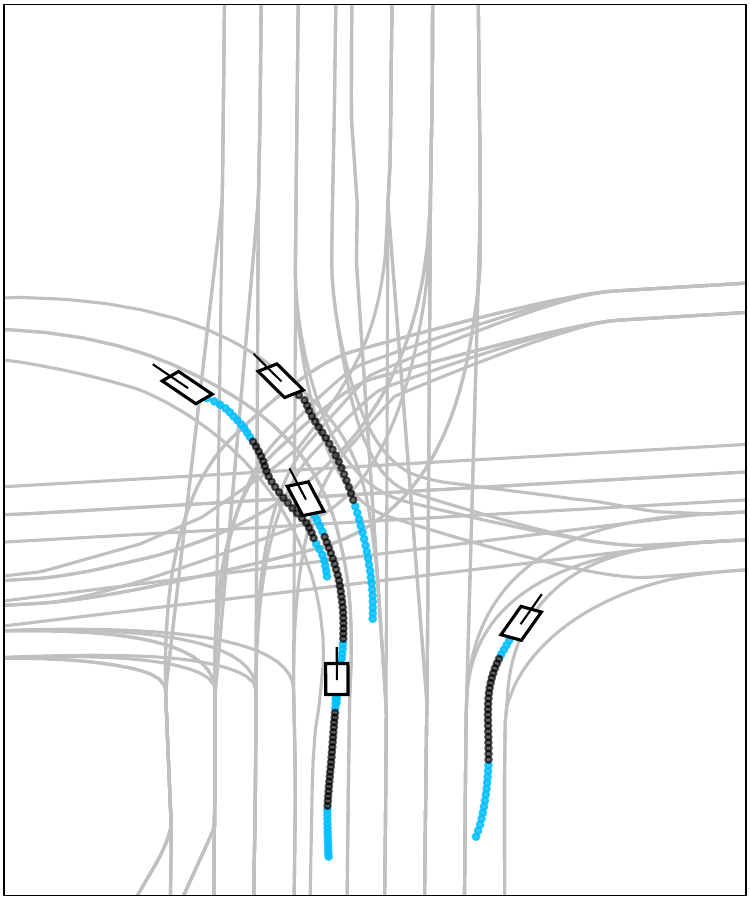}
    \caption{MTR}
    \label{fig_ssl1}
  \end{subfigure}
  \hspace{0.5pt}
  \begin{subfigure}[b]{0.22\textwidth}
    \includegraphics[width=\textwidth]{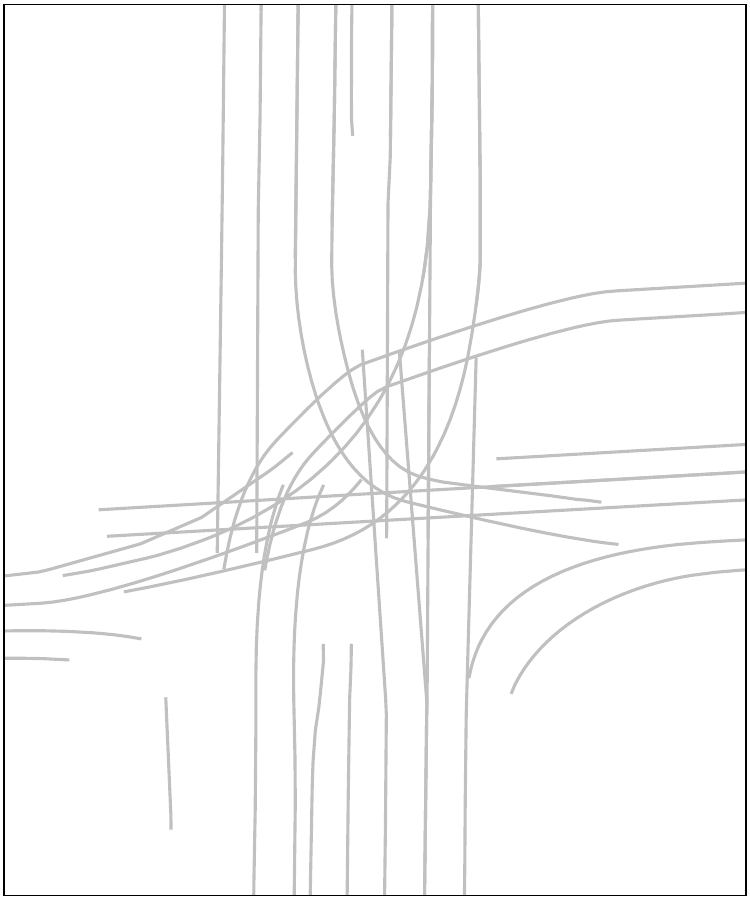}
    \caption{MRR}
    \label{fig_ssl2}
  \end{subfigure}
  \hspace{0.5pt}
  \begin{subfigure}[b]{0.22\textwidth}
    \includegraphics[width=\textwidth,trim={0cm 0cm 0cm 0cm},clip]{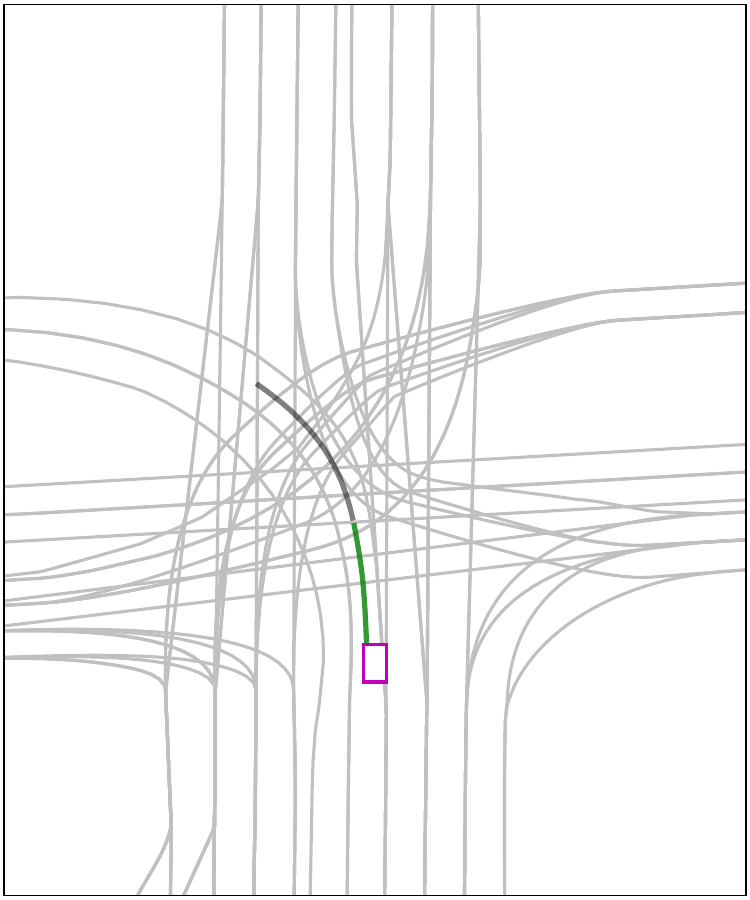}
    \caption{MNR}
    \label{fig_ssl3}
  \end{subfigure}
  \hspace{0.5pt}
  \begin{subfigure}[b]{0.22\textwidth}
    \includegraphics[width=\textwidth,trim={0cm 0cm 0cm 0cm},clip]{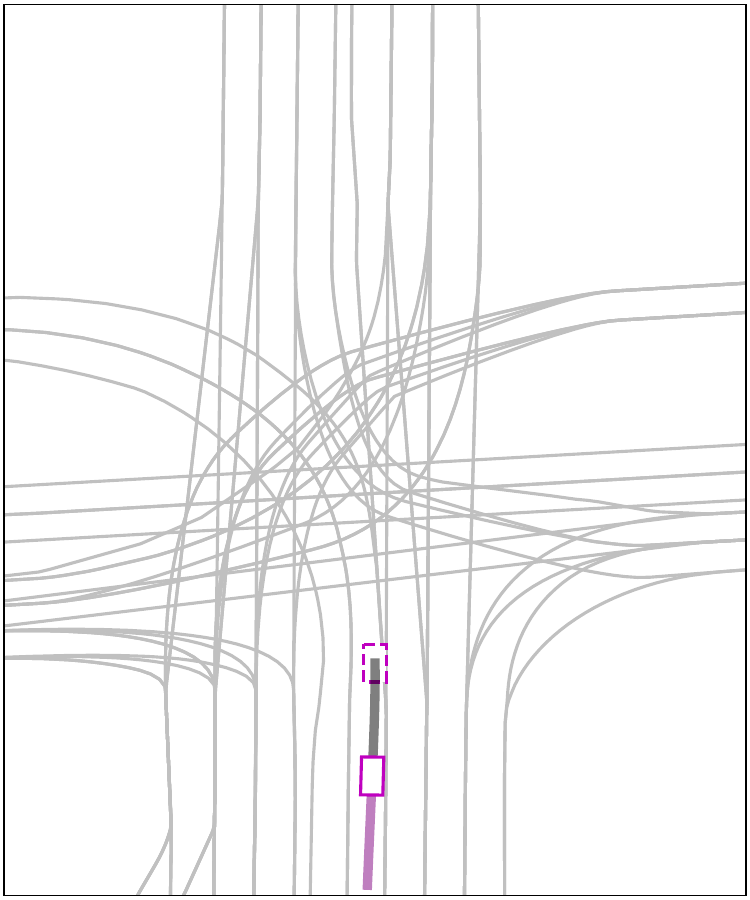}
    \caption{ETR}
    \label{fig_ssl4}
  \end{subfigure}
  \caption{Illustration of the subtasks in Plan-MAE: (a) masked agent trajectories (black) with the remaining portion (blue); (b) map segments after random masking;(c) masked navigation (black) with the remaining portion (green); (d) head (red) and tail (black) trajectory of the ego.}
  \label{fig_ssl_tasks}
\end{figure}

\textbf{Masked Trajectory Reconstruction (MTR)} enhances the capacity of the agent encoder to model agents' motion patterns by randomly masking a portion of their trajectories along the temporal axis. To ensure informative learning, the padded virtual agents are excluded from the masking targets, and those with trajectory lengths shorter than $15$m are also ignored. For the remaining agents, their trajectories are randomly masked at a 0.5 ratio, i.e., a mask length of $T_h/2$ for trajectories spanning $T_h$ frames, as shown in Figure~\ref{fig_ssl1}. These masked trajectories are fed into the agent encoder, and a two-layer MLP is used to reconstruct the missing frames. 

\textbf{Masked Road Reconstruction (MRR)} focuses on the spatial understanding of road topology for the map encoder. As shown in Figure~\ref{fig_ssl2}, a portion of map lanes is randomly masked and then reconstructed using the surrounding map context and relative poses. This forces the map encoder to capture topological continuity and geometric alignment among road elements. To prevent introducing invalid features, we do not mask padding lanes or those shorter than $20$m. Besides, considering that the head and tail of excessively long lanes are far from the ego vehicle, we randomly mask a contiguous segment of 100 points within the middle 200 points of each valid lane, i.e., the mask ratio is 0.5.

\textbf{Masked Navigation Reconstruction (MNR)} introduces the goal-aware capability by masking segments of the navigation route, and enables the navigation encoder to reconstruct the missing components. Figure~\ref{fig_ssl3} depicts a scenario that the ego vehicle should turn left at this crossroad, wherein the navigation reconstruction can enhance the understanding of riding direction and road structures. This is essential for motion planning as it must generate trajectories that strictly conform to navigation destinations. Since the navigation route extends more than 1500m ahead of the ego vehicle, we only input 20 points close to the vehicle into the model and use a two-layer MLP network to predict the next 20 points.

\textbf{Ego Tail Reconstruction (ETR)} is the alignment task in Plan-MAE, integrating the auxiliary tasks to achieve pretraining of the entire encoder. As shown in Figure~\ref{fig_ssl4}, it evenly divides $T_h$-frame trajectory of the ego into the head and tail segments, and predicts the tail from the head. In this case, the current frame is regressed back from $T_h+1$ to $T_h/2+1$ and serves as the input to the ego-state encoder. Differently, this local planning employs complete embeddings of map lanes, agents and navigation information, rather than their masked counterparts to generate the tail segment. Similar to the train from scratch manner, the fusion encoder is used to aggregate the embeddings, and the context is delivered to a two-layer MLP to output the ego trajectory.

\subsection{Finetune for Prediction and Planning}
In the finetune stage, we integrate the pretrained encoder with the decoder, and train the entire model to learn the prediction and planning via supervised learning. The training, validation, and test datasets are maintained consistently with those used in the pretraining stage, where their labels are annotated as the future trajectories of the ego vehicle and agents. For the training loss, we combine the widely used trajectory regression loss and the confidence classification loss with the equal weights. For the $i$-th agent, the regression loss is calculated by the $L_2$-distance between the best predicted trajectory $\tau^{*}_i$ and the ground truth $\tau_{ig}$. And the classification loss is computed by the categorical cross entropy of the output scores against the one-hot index of the best trajectory. Considering the ego vehicle and $N$ agents present with $m$ trajectories, the finetune loss can be defined as:
{\allowdisplaybreaks
    \begin{align*}
        \mathcal{L} &=\sum_{i=1}^{N+1} {\mathcal{L}_{\rm traj}^i + \mathcal{L}_{\rm score}^i},\\
        \mathcal{L}_{\rm traj}^i &= \left\lVert \tau^{*}_i - \tau_{ig} \right\rVert_2,\\
        \mathcal{L}_{\rm score}^i &= -\sum_{j=1}^{m} \big\{\mathbb{I}_{(j={ig})}\log p_j  + \mathbb{I}_{(j \neq {ig})}(1-\log p_j)\big\}.
    \end{align*}
}
where $\mathbb{I}$ is the binary indicator function.

\section{Experiments}
\subsection{Experimental Setup}

\paragraph{Dataset.} We use production-ready vehicles to collect expert driving data in Chongqing and Shanghai, China. Totally, we gather 3.1 million driving segments for training, and 31 thousand segments for validation and testing. Each sample contains 4 seconds of past and 5 seconds of future trajectory, with the time interval between adjacent frames as 0.1 seconds, i.e., $T_h=40$ and $T_f=50$. The instance number $N$ and $M$ are set as 30, the point number $P_m$ and $P_r$ are set as 200, 500 receptively, the attribute numbers are set as $D_e=11, D_n=13, D_m=8, D_r=3$.

\paragraph{Metrics.} {Following the widely used evaluation metrics in the planning, we employ the average displacement error (ADE) and final displacement error (FDE) of the best planned trajectory and the most confident trajectory, to formulate the minADE, minFDE, topADE, topFDE respectively. Here, the best trajectory refers to the one with the smallest average error relative to the ground-truth, and the most confident trajectory refers to the one with the highest score. Besides, the collision rate is introduced to evaluate the safety of planning, defined as the ratio of the best planned trajectory that collides with the ground-truth of agents. Lower values across all metrics suggest better alignment with ground-truth trajectories.}

\paragraph{Baselines.}
We first consider 3 strong baselines that directly train the model with an imitative manner:
(1) \textbf{Vanilla IL} aims to directly predict the ego’s future trajectory, reuses the encoder from our model, but simplifies the decoder to a single layer. The output is a single planned trajectory.
(2) 
\textbf{GameFormer}~\cite{huang2023gameformer} employs a similar Transformer-based architecture to our model, but lacks the state dropout encoder and corresponding state perturbation, maintaining consistency with the original paper's configuration. 
(3)
\textbf{PlanTF}~\cite{cheng2024rethinking} introduces a state dropout encoder to promote robust feature usage and applies trajectory perturbation with feature normalization to reduce compounding errors. Besides, the prediction pretraining methods, \textbf{Traj-MAE}~\cite{chen2023traj}, \textbf{Forecast-MAE}~\cite{cheng2023forecast} and \textbf{SEPT}~\cite{lan2023sept}, are adapted to incorporate the ego's trajectory by treating it as an ordinary agent, without accounting for its navigation route.

\paragraph{Implementation details.}
Both the pretraining and finetune stages are trained on 40 NVIDIA H20 GPUs, with a batch size of 32 per GPU. And they also use the same training, validation and testing datasets while the data labels are dropped during pretraining. Besides, the model is trained for 32 epochs for pretraining and 24 epochs for finetune. We first use a linear learning rate for warm-up and then a cosine annealing one which decays progressively over epochs from $1 \times 10^{-4} $ down near 0. The embedding dimension $D$ is 256, the output modality $m$ is 12, the $K_T$ and $K_C$ of the model are both 6.

\subsection{Experimental Results}
The numerical results on the prediction and planning metrics are listed in Table \ref{tab:baselines}. Plan-MAE achieves the best performance across the planning distance metrics and the lowest collision rate, meanwhile its prediction performance is comparable with other pretraining methods.
Compared to Gameformer trained from scratch, Plan-MAE reduces minADE and minFDE of planning by 39.60\% and 37.24\%, respectively. This means that the introduction of self-supervised pretraining can enhance the spatiotemporal understanding of traffic elements, which provides effective representations for the ultimate motion planning. Compared to the best pretraining method SEPT, Plan-MAE shows a clear advantage in planning metrics, reducing topADE and topFDE by 7.89\% and 9.03\%, respectively. This indicates that although both prediction and planning involve trajectory forecasting, the dedicated ego-centric tasks, MNR and ETR, can provide more significant improvements for planning tasks.

Note that GameFormer achieves the best prediction in Table \ref{tab:baselines}. This is because all the pretraining methods and PlanTF introduce data augmentation, which can improve planning performance but suppress prediction. As the gains in prediction and planning do not always align, planning-oriented optimization is provably a more reasonable approach to enhance the overall driving performance.

\begin{table*}[!htbp]
\caption{Comparison with the baselines on test dataset.}
\centering
\label{tab:baselines}
\begin{tabular}{l|cc|cccc|c}
\toprule
\multirow{2}{*}{Method} & \multicolumn{2}{c|}{Prediction @5s} & \multicolumn{4}{c|}{Planning @ 5s} & \multirow{2}{*}{\makecell{Collision \\ rate (\%)}}\\
\cmidrule(r){2-7}
                       & minADE & minFDE & minADE &  minFDE & topADE & topFDE \\
\midrule
Vanilla IL         & - & - & 1.933 & 3.886 & 1.933 & 3.886 &  0.127 \\
GameFormer         & \textbf{0.933} & \textbf{1.770} & 0.884 & 1.262 & 3.418 & 6.992 & 0.116\\
PlanTF             & 0.961 & 1.848 & 0.574 & 0.804  & 1.174 & 2.906 & 0.121\\
\midrule
Traj-MAE           & 0.950 & 1.936 & 0.568 & 0.824 & 1.082 & 2.641 & 0.123\\
Forecast-MAE       & 0.983 & 1.835 & 0.568 & 0.848  & 1.164 & 2.835 & 0.119\\
SEPT               & 0.948 & 1.826 & 0.558 & 0.818 & 1.115 & 2.834  & 0.118 \\
Plan-MAE           & 0.964 & 1.817 & \textbf{0.534} & \textbf{0.792} & \textbf{1.027} & \textbf{2.578} & \textbf{0.112}\\
\bottomrule
\end{tabular}
\end{table*}

\subsection{Visualizations}

\begin{figure}[ht]
    \centering
    \begin{subfigure}[b]{0.325\linewidth}
        \includegraphics[width=\textwidth]{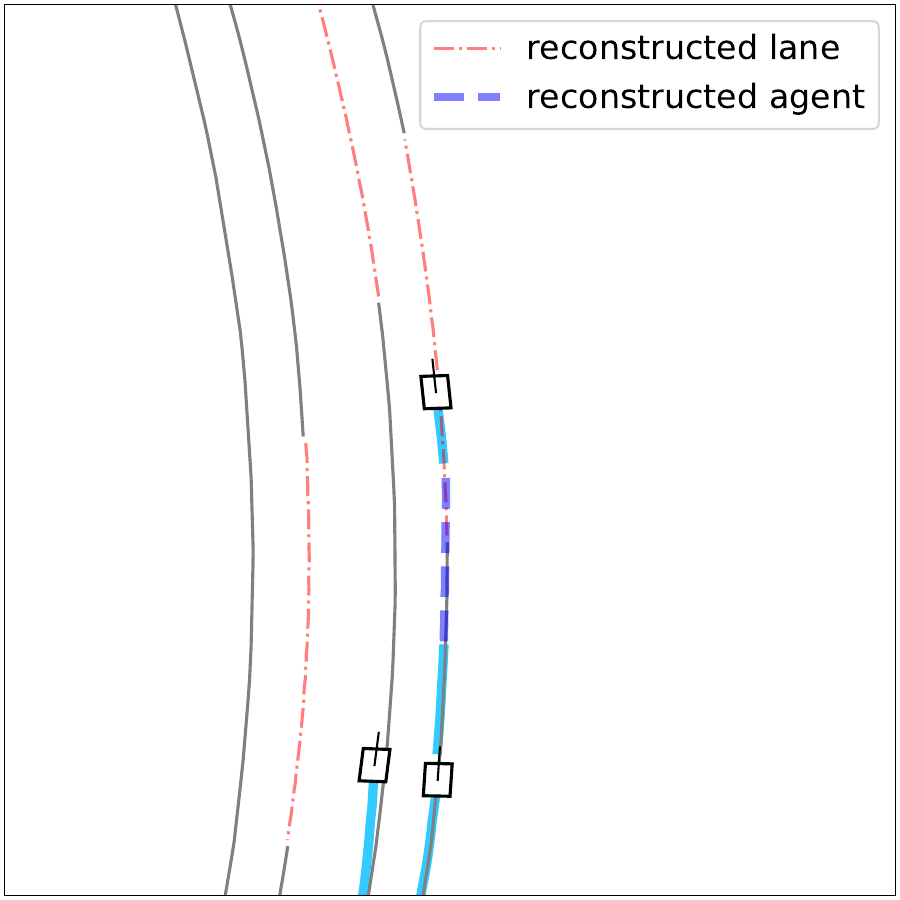}
        \caption*{}
    \end{subfigure}
    \hfill
    \begin{subfigure}[b]{0.325\linewidth}
        \includegraphics[width=\textwidth]{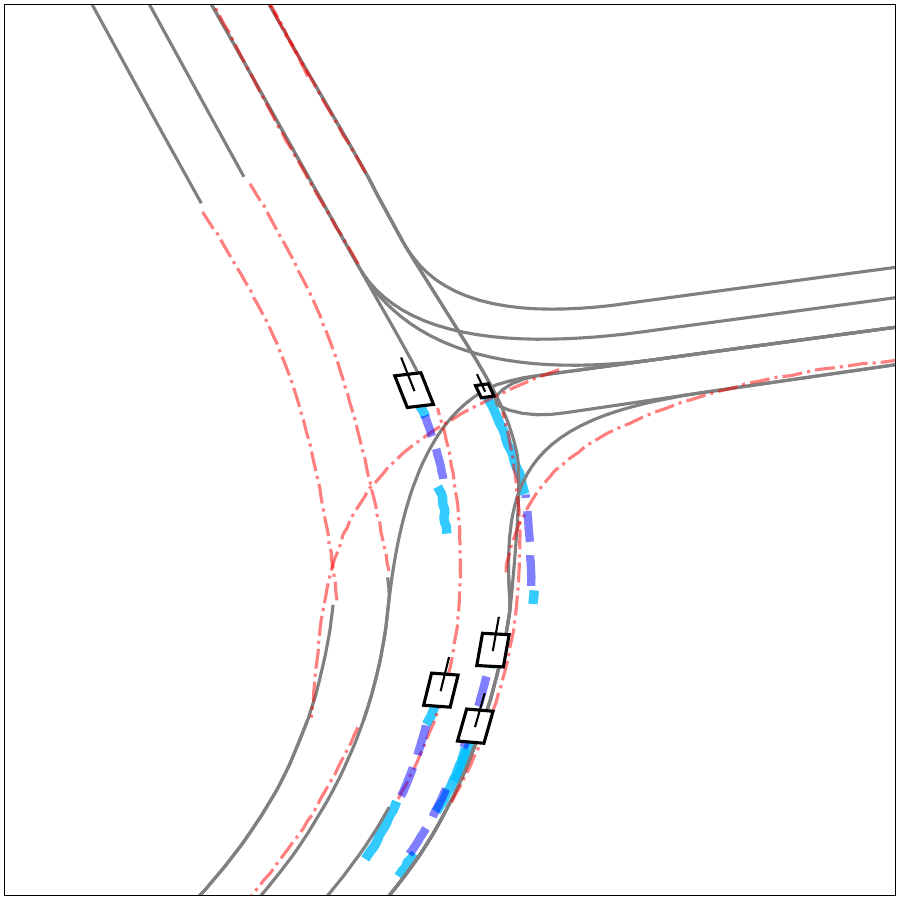}
        \caption*{}
    \end{subfigure}
    \hfill
    \begin{subfigure}[b]{0.325\linewidth}
        \includegraphics[width=\textwidth]{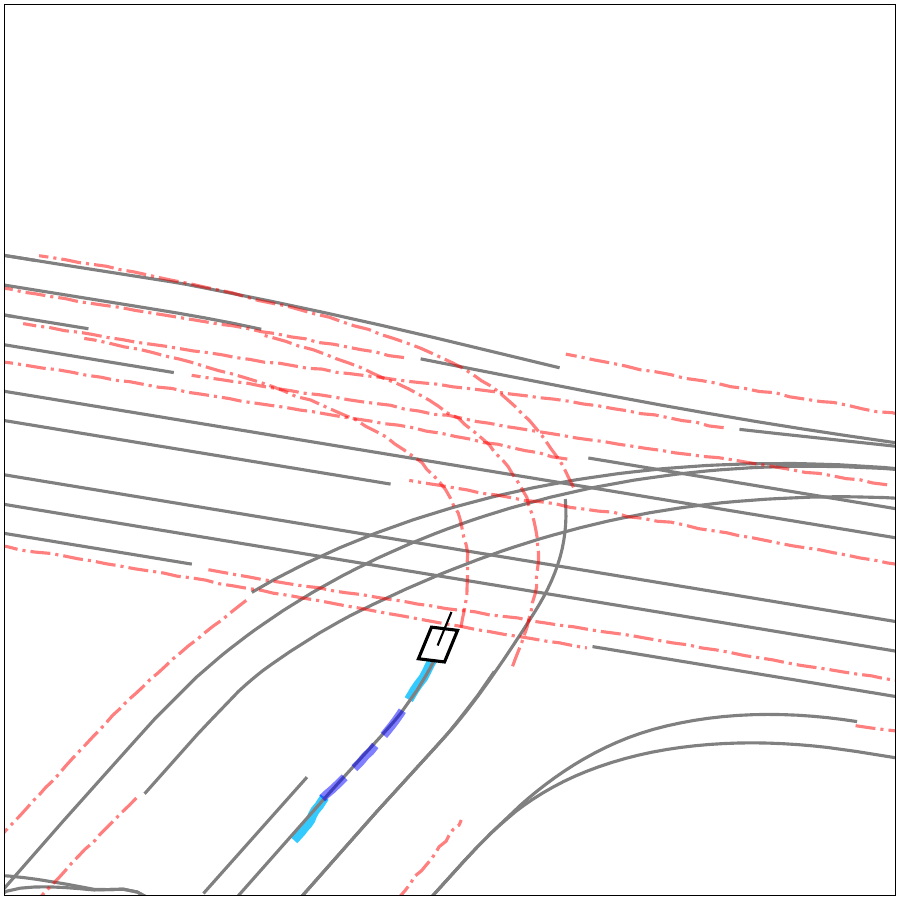}
        \caption*{}
    \end{subfigure}

    \vspace{-1.5em} 

    \begin{subfigure}[b]{0.325\linewidth}
        \includegraphics[width=\textwidth]{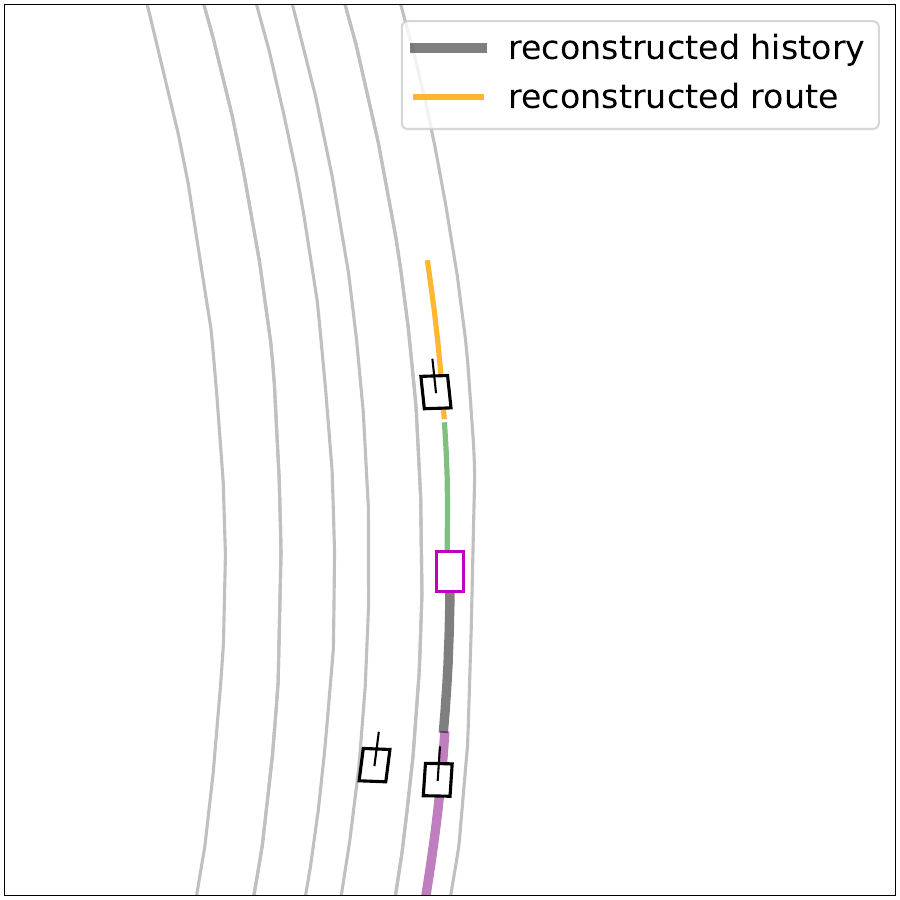}
        \caption{case 1\label{fig:recons_vis1}}
    \end{subfigure}
    \hfill
    \begin{subfigure}[b]{0.325\linewidth}
        \includegraphics[width=\textwidth]{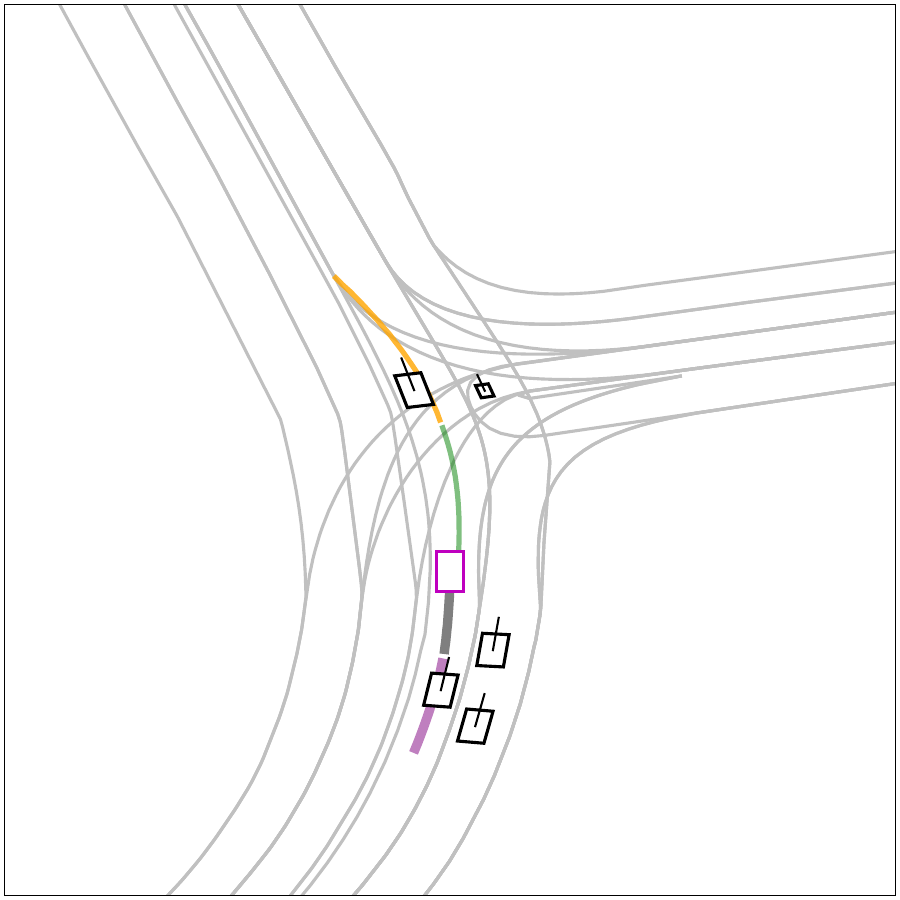}
        \caption{case 2\label{fig:recons_vis2}}
    \end{subfigure}
    \hfill
    \begin{subfigure}[b]{0.325\linewidth}
        \includegraphics[width=\textwidth]{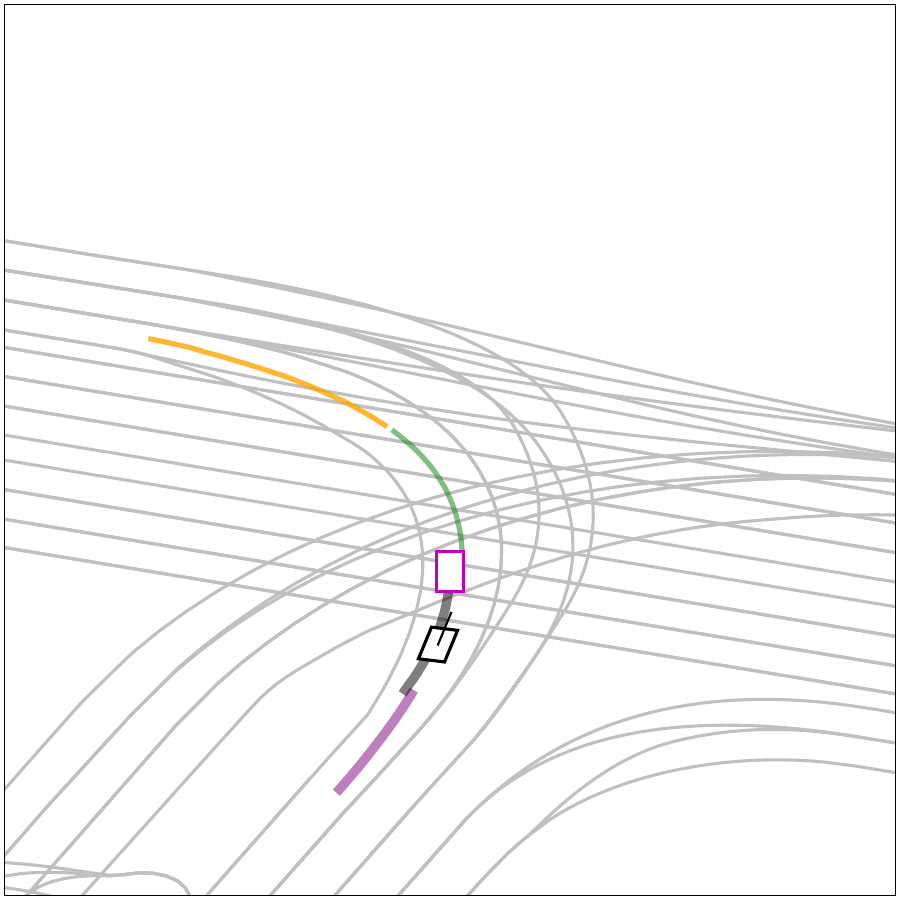}
        \caption{case 3\label{fig:recons_vis3}}
    \end{subfigure}

    \caption{Reconstruction visualization of the pretraining tasks on three cases. 
    The ego vehicle and surrounding agents are marked by magenta and black rectangles, respectively.  
    The top row shows outputs from MTR and MRR, while the bottom row corresponds to the ego-centric MNR and ETR.}
    \label{fig:recons_vis}
\end{figure}

\paragraph{Pretraining reconstruction.} Figure~\ref{fig:recons_vis} presents reconstruction outcomes across all subtasks following pretraining. The top row demonstrates the scene-centric MTR and MRR, which accurately reconstruct both road geometries and multi-agent motion patterns. These reconstructions capture fine-grained map topologies and socially compliant trajectories, indicating robust comprehension of spatial semantics and interactive dynamics. The bottom row displays the ego-centric MNR and ETR, where the model faithfully predicts intent-aware navigation routes and reconstructs the ego vehicle's short-term motion. The resulting trajectories exhibit smoothness, kinematic feasibility, and strong congruence with ground-truth motion. Collectively, these tasks provide complementary learning signals: scene-centric reconstructions strengthen contextual understanding, while ego-centric reconstructions guide physically constrained and goal-oriented planning representations in Plan-MAE.

\begin{figure}[ht]
    \centering
    \begin{subfigure}[b]{0.325\linewidth}
    \includegraphics[width=\textwidth]{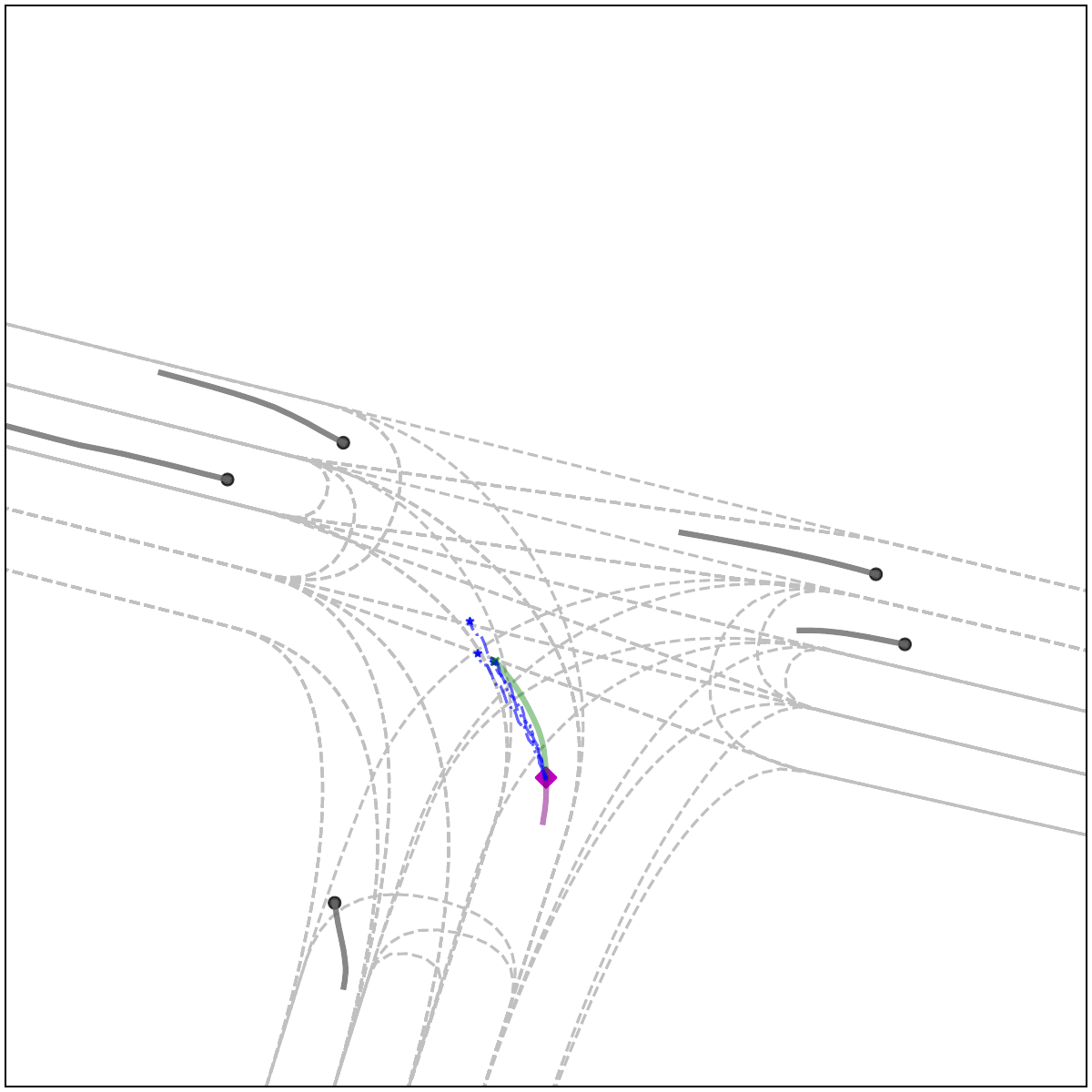}
        \caption*{}
    \end{subfigure}
    \hfill
    \begin{subfigure}[b]{0.325\linewidth}
        \includegraphics[width=\textwidth]{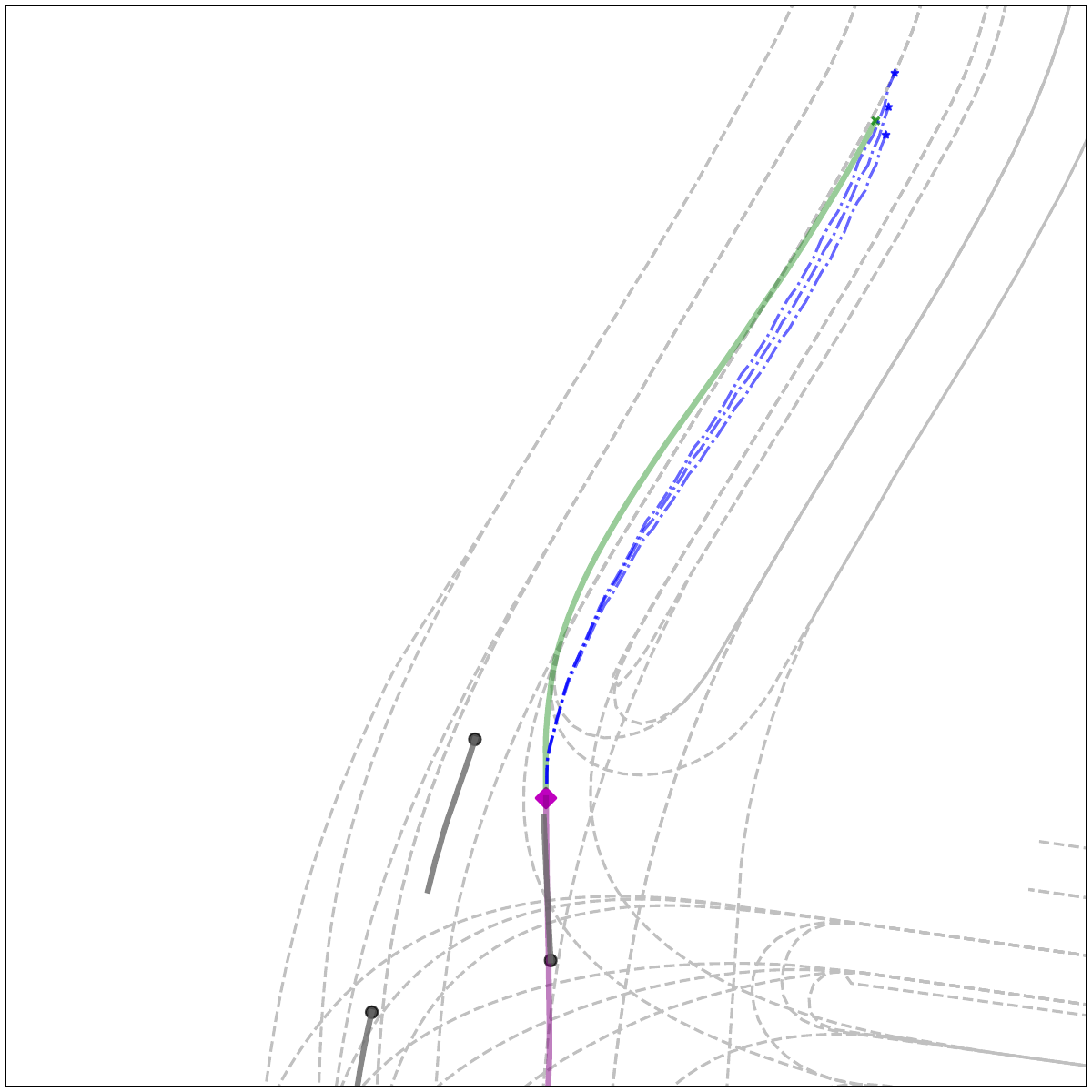}
        \caption*{}
    \end{subfigure}
    \hfill
    \begin{subfigure}[b]{0.325\linewidth}
        \includegraphics[width=\textwidth]{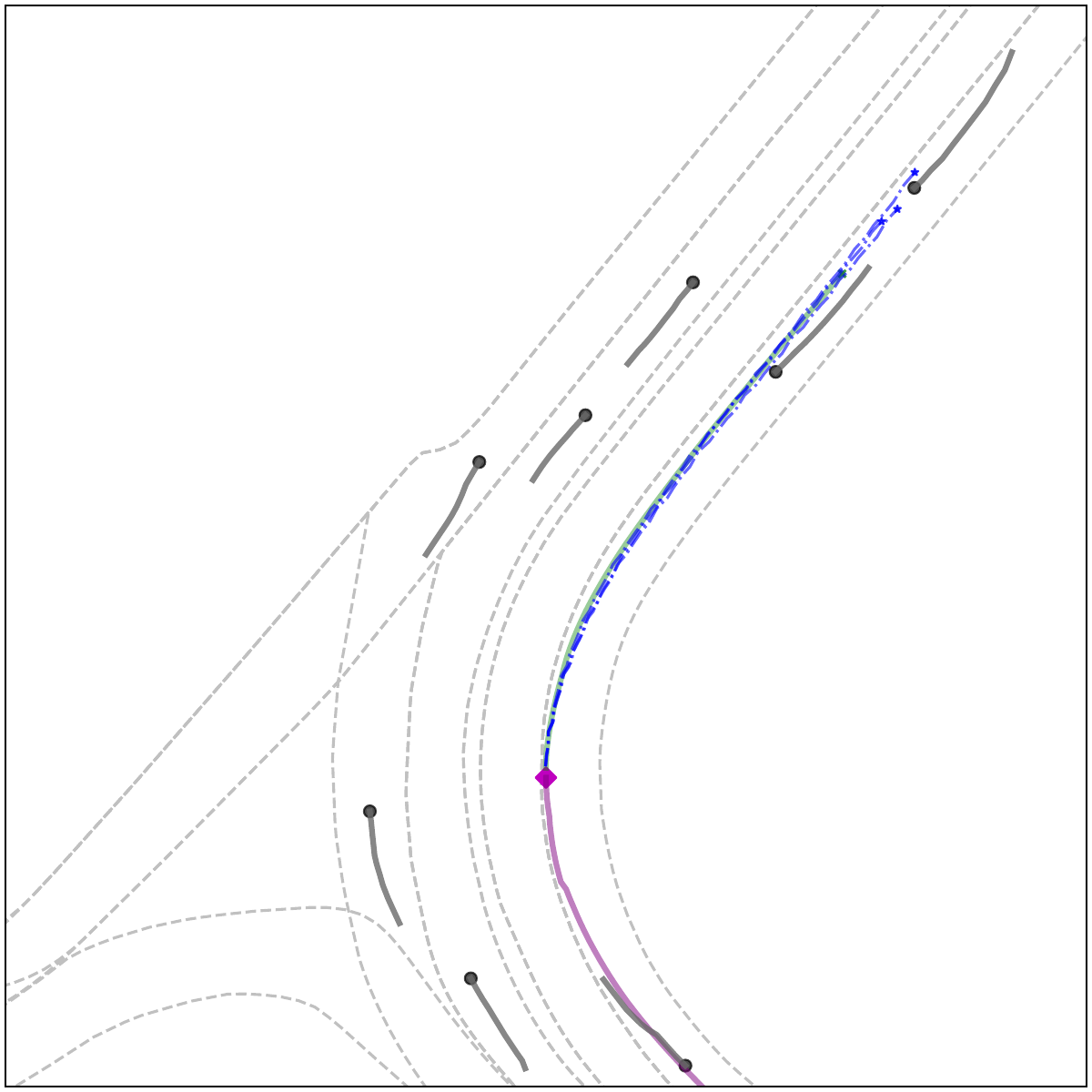}
        \caption*{}
    \end{subfigure}

    \vspace{-1.5em} 

    \begin{subfigure}[b]{0.325\linewidth}
        \includegraphics[width=\textwidth]{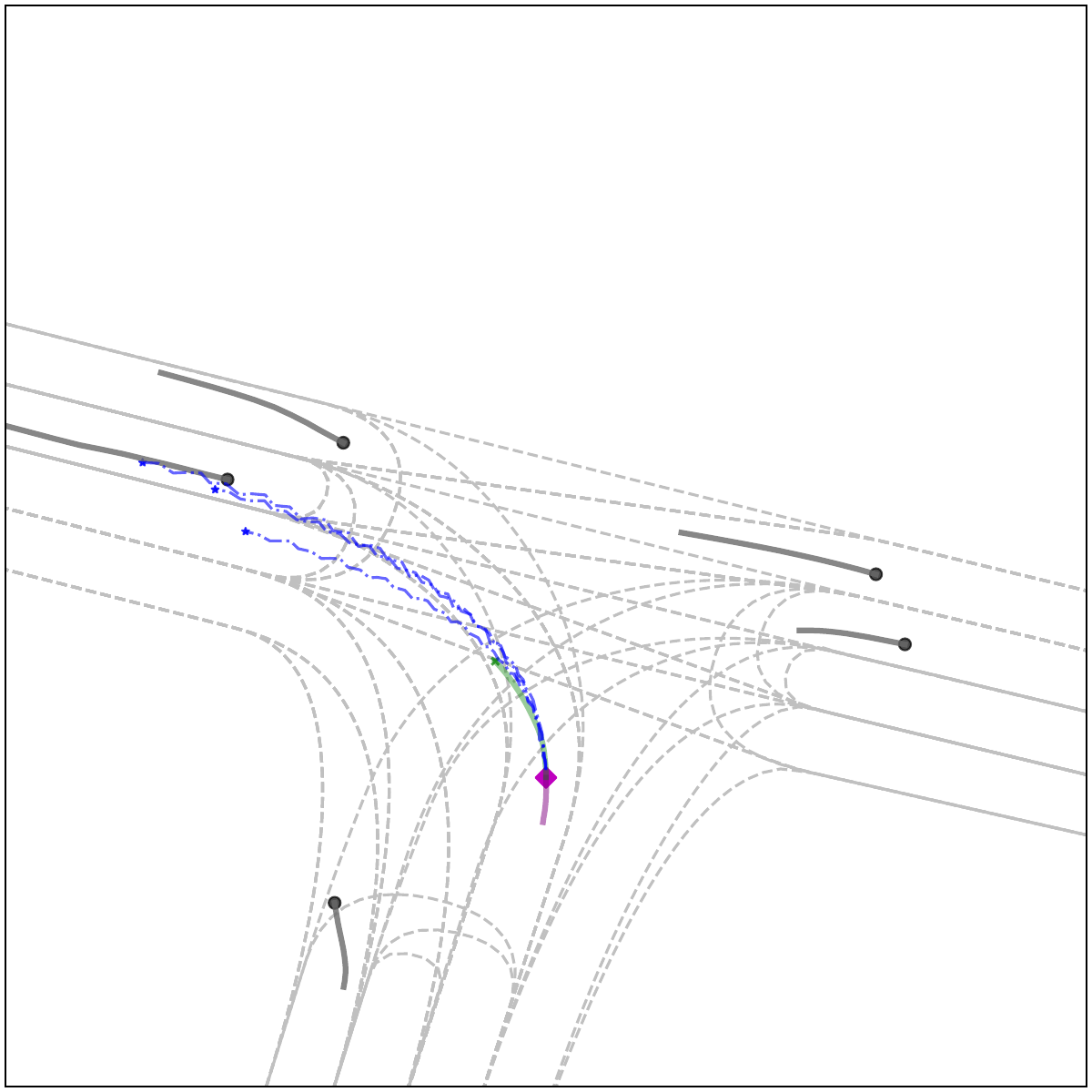}
        \caption{case 1\label{fig:vis_case1}}
    \end{subfigure}
    \hfill
    \begin{subfigure}[b]{0.325\linewidth}
        \includegraphics[width=\textwidth]{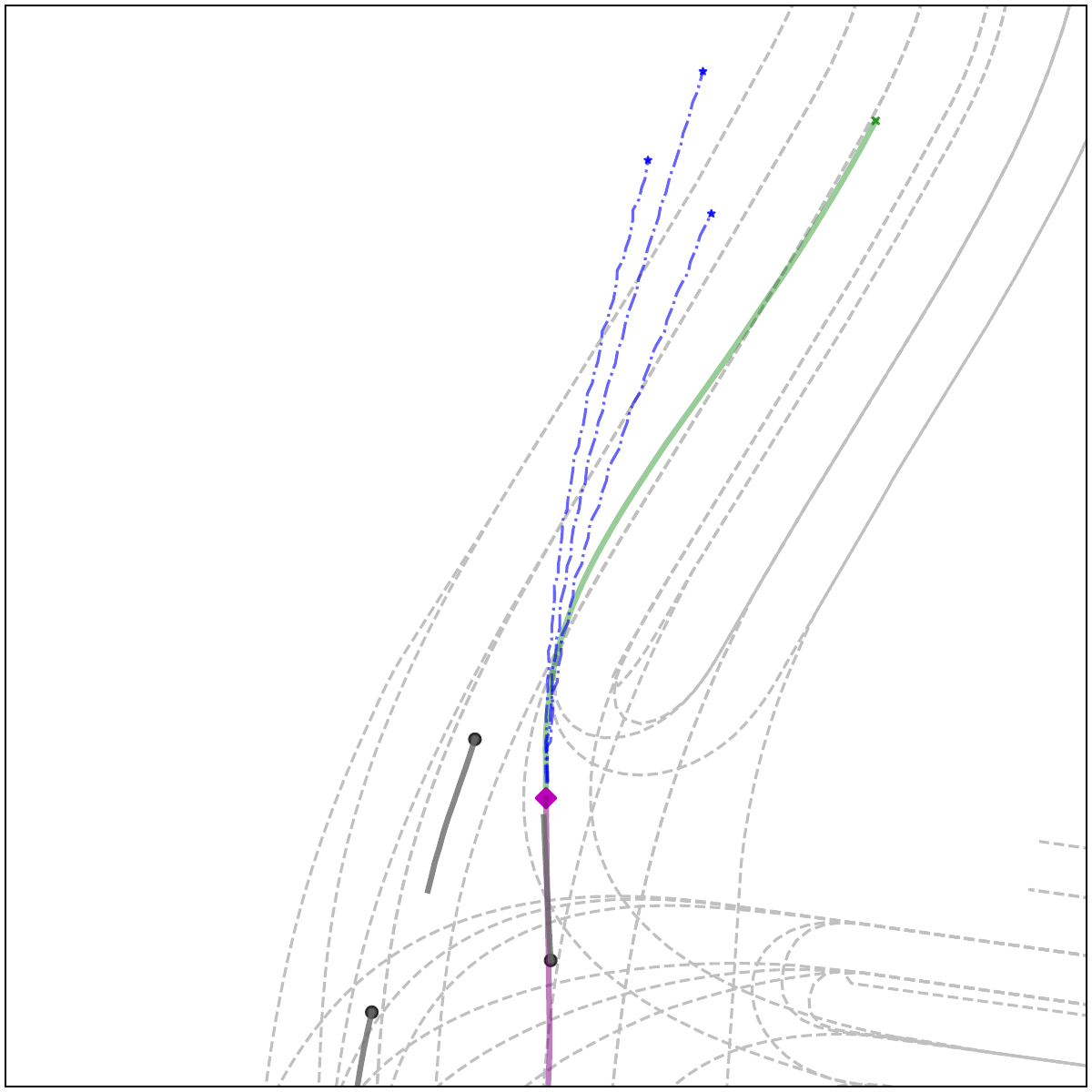}
        \caption{case 2\label{fig:vis_case2}}
    \end{subfigure}
    \hfill
    \begin{subfigure}[b]{0.325\linewidth}
        \includegraphics[width=\textwidth]{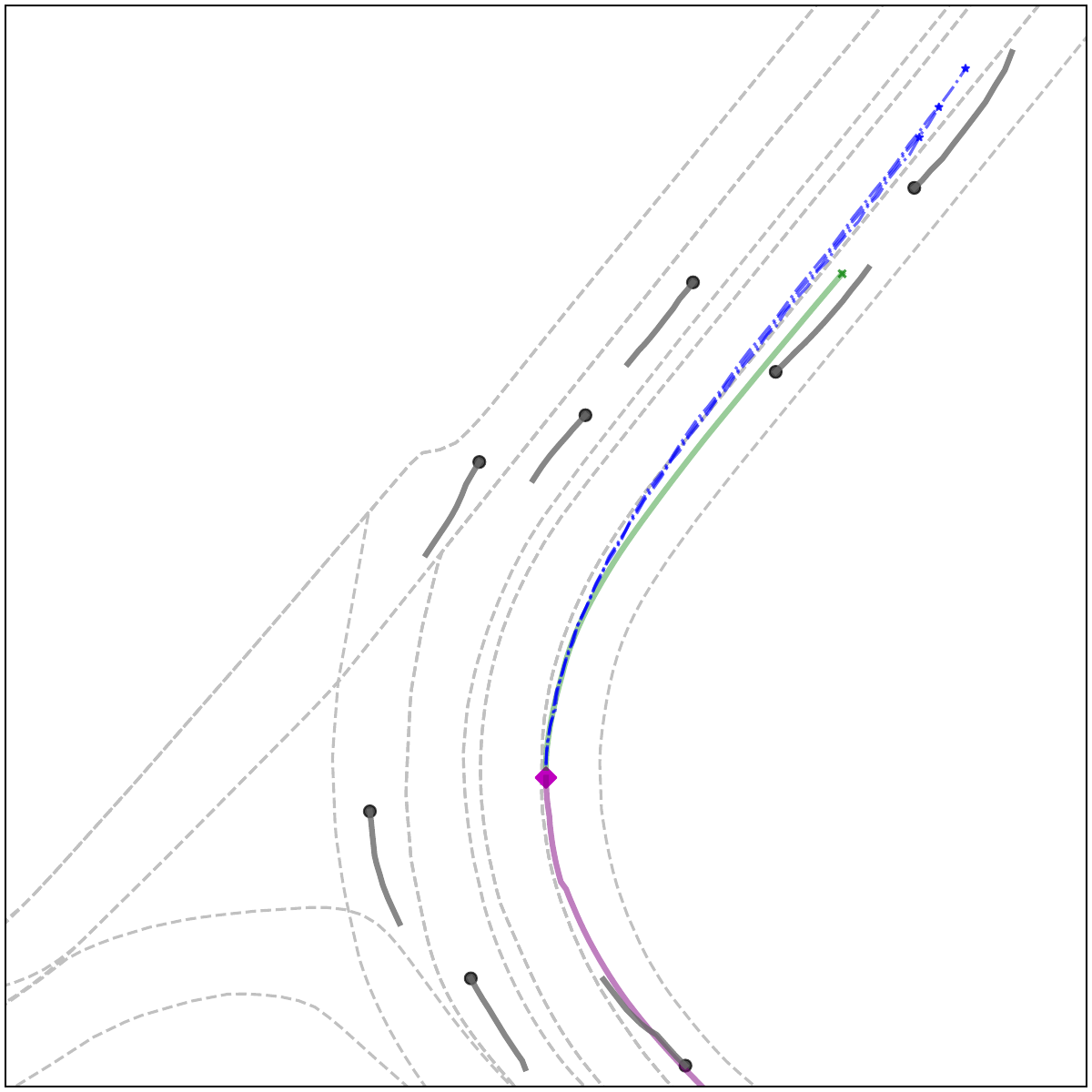}
        \caption{case 3\label{fig:vis_case3}}
    \end{subfigure}

    \caption{Visualization of three driving scenarios. The ego vehicle is represented by a red diamond, with its history and future ground truth represented by red and green curves, respectively. The top-3 planned trajectories are shown as blue curves. In each case, the upper image is inferenced by the model with the plan-MAE pretraining, while the lower image corresponds to the model that learns from scratch.}
    \label{fig:plan_vis}
\end{figure}

\paragraph{Motion planning.}
To better understand the impact of pretraining on planning, we provide qualitative comparisons in Figure~\ref{fig:plan_vis} between our models and its counterpart learning from scratch. Across all examples, the trajectories generated by Plan-MAE are visibly closer to the ground-truth future motion, while the models without pretraining exhibit abrupt accelerations or behaviors that violate road constraints. In Figure~\ref{fig:vis_case1}, plan-MAE enables the ego vehicle to slow down at a T-junction, yielding to the vehicle going straight ahead, as shown by the ground-truth from human drivers. In contrast, the model without pretraining would accelerate directly, violating the traffic-light restrictions at the intersection. In Figure~\ref{fig:vis_case2}, the pre-training enables the planning model to produce trajectories that conform to the current road curvature constraints, whereas the model trained from scratch plans infeasible trajectories that deviate from the lane. In Figure~\ref{fig:vis_case3}, when the ego vehicle is turning right near the lane line, the planned trajectories of the pretrained model gradually move toward the lane center, whereas the model trained from scratch plans trajectories that remain on the line.
\begin{figure*}[ht]
  \centering
  \begin{subfigure}[b]{0.32\textwidth}
    \includegraphics[width=\textwidth]{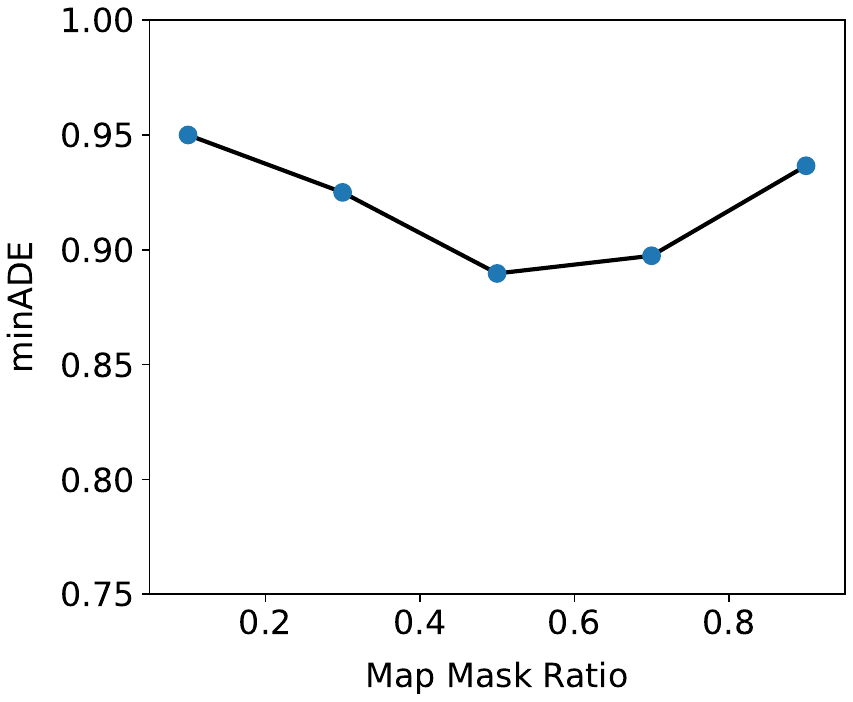}
    \caption{MRR}
    \label{fig:ratio_map}
  \end{subfigure}
  \hspace{0.1pt}
  \begin{subfigure}[b]{0.32\textwidth}
    \includegraphics[width=\textwidth,trim={0cm 0cm 0cm 0cm},clip]{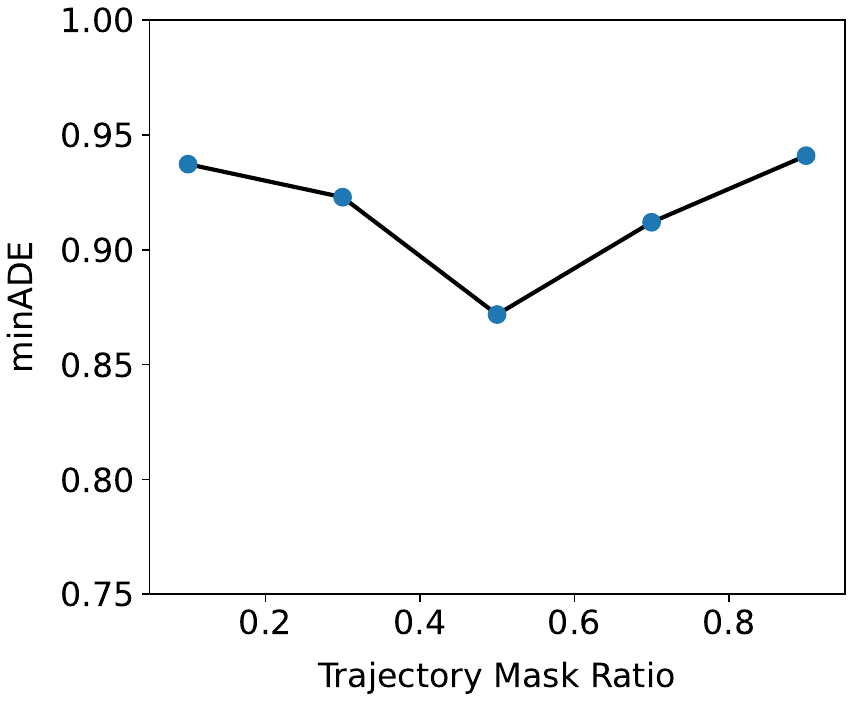}
    \caption{MTR}
    \label{fig:ratio_hist}
  \end{subfigure}
  \hspace{0.1pt}
  \begin{subfigure}[b]{0.32\textwidth}
    \includegraphics[width=\textwidth,trim={0cm 0cm 0cm 0cm},clip]{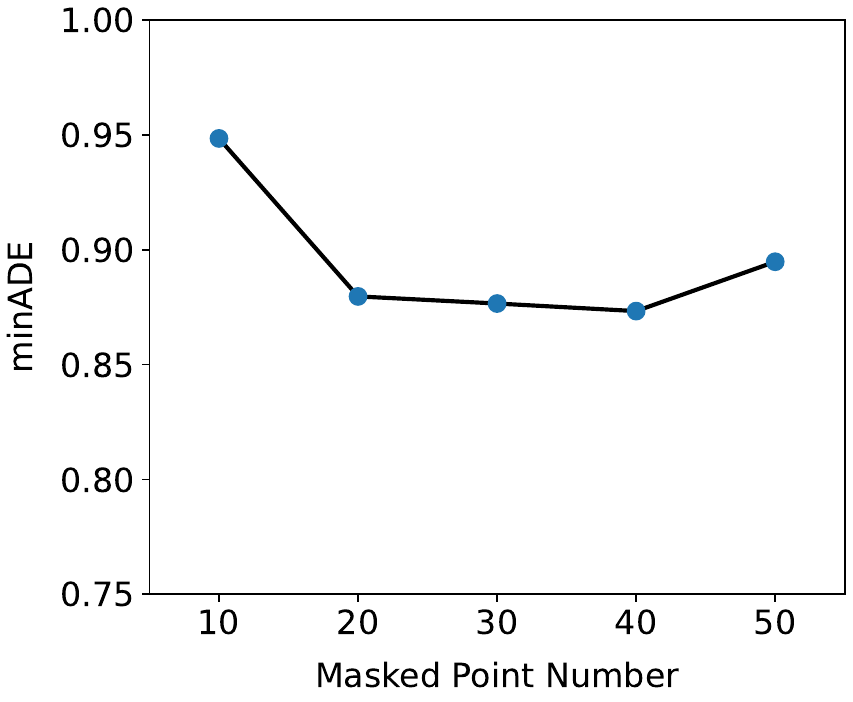}
    \caption{MNR}
    \label{fig:ratio_route}
  \end{subfigure}
  \caption{Impact of mask length in the three auxiliary tasks.}
  \label{fig:mask_ratio_effect}
\end{figure*}

\subsection{Ablation Study}
\paragraph{Effectiveness of each sub-task in the pretraining.}
To disentangle the contributions of each self-supervised objective in Plan-MAE, we conduct a systematic ablation by selectively removing one sub-task at a time. Besides, we explore the influence of alignment task by replacing ETR with MNR as the ultimate procedure. In this case, the model prioritizes deriving future trajectories based on navigation route ahead of the ego vehicle, rather than relying on historical trajectories. The results on the planning metrics are shown in Table~\ref{tab_ablation1}.

Collectively, ablating any sub-task in Plan-MAE degrades planning performance, confirming that all four sub-tasks contribute positively and complementarily to the final planning objective. Notably, MNR exhibits the most pronounced impact on the minADE and minFDE, underscoring that the directional guidance from navigation is helpful to generate accurate trajectories. ETR exerts the most significant impact on the topADE and topFDE, indicating the information such as vehicle kinematics and obstacle avoidance from the ego trajectory is beneficial to the score head to identify a reasonable trajectory. Meanwhile, removing MTR results in a slight reduction in minADE, which is due to the fact that perceived agents are subject to significant noise, making certain segments difficult to reconstruct. However, it still leads to a noticeable increase in topFDE, highlighting that capturing social interaction is critical for robust planning.
As shown in the last row of Table~\ref{tab_ablation1}, using MNR as the alignment results in the highest minFDE and topFDE, suggesting ETR is a better choice for alignment task. As it leverages historical trajectories recoded from human drivers, ETR can provide more informative and physically grounded supervision to the downstream planning.

\begin{table}[!htbp]
\caption{Effectiveness of each sub-task to planning metrics.}
\centering
\footnotesize
\label{tab_ablation1}
\begin{tabular}{lcccc}
\toprule
Variant & minADE & minFDE & topADE & topFDE \\
\midrule
Plan-MAE & 0.534 & \textbf{0.792} & \textbf{1.027} & \textbf{2.578} \\
w/o MRR & 0.545  & 0.814  & 1.096  &  2.662 \\
w/o MTR & \textbf{0.527}  & 0.805  & 1.067  & 2.691   \\
w/o MNR & 0.570 & 0.825 & 1.090 & 2.601 \\
w/o ETR & 0.547 & 0.798  & 1.110 & 2.732 \\
\midrule
ETR $\rightarrow$ MNR & 0.565 & 0.833 & 1.163 & 2.921 \\
\bottomrule
\end{tabular}
\end{table}

\paragraph{Impact of mask length.}
We further investigate the impact of mask length in auxiliary tasks on planning performance. For MTR and MRR, the lengths of masked trajectory or lane are governed by the mask ratio, whereas in MNR, mask length corresponds to the number of waypoints masked along the navigation route.
Figure~\ref{fig:mask_ratio_effect} demonstrates the trend of planning minADE with respect to varying mask lengths. In Figure~\ref{fig:ratio_map},  the planning performance initially improves and then degrades as the map mask ratio increases from 0.1 to 0.9. This indicates that insufficient masking limits contextual learning, while excessive masking compounds the difficulty of reconstructing. Figure~\ref{fig:ratio_hist} 
reveals a parallel phenomenon when the trajectory mask ratio increases. Consequently, we uniformly apply a 0.5 mask ratio to both map and trajectory inputs in the Plan-MAE. Figure~\ref{fig:ratio_route} explores the influence of masked point numbers (from 10 to 50) on the MNR. Clearly, there exists a broad intermediate range of mask settings where the performance remains consistently strong, suggesting that our method is not overly sensitive to the navigation length. We choose to mask the closest 20 points against the ego and prediction the next 20 points in our experiment.

\section{Conclusion}
In this paper, we present Plan-MAE, a self-supervised pretraining framework for the integrated prediction and planning of autonomous driving that capitalizes on the masked autoencoder. By designing three auxiliary tasks which focus on the agent trajectory, road geometry and navigation route, our approach enables the model to learn rich scene representations from reconstructing the masking elements. Besides, the alignment task is utilized to integrate features from the auxiliary tasks by conducting a short-term planning with the ego's history. Extensive experiments on the real-world dataset demonstrate that Plan-MAE achieves superior planning accuracy and safety over existing self-supervised and supervised baselines. Ablation studies further highlight the importance of each self-supervised objective in building a comprehensive understanding of driving scenarios. We believe Plan-MAE establishes a scalable foundation for motion planning while providing architectural insights for integrating structured scene cognition into data-driven driving systems.

\bibliographystyle{plain}
\bibliography{neurips_2025}

\newpage
\appendix








\end{document}